\documentclass[letterpaper, 10 pt, conference]{ieeeconf}  % Comment this line out if you need a4paper

\IEEEoverridecommandlockouts                              % This command is only needed if 
\usepackage{amsmath}
\usepackage{cite}
\usepackage{graphicx}
\usepackage[ruled,linesnumbered,vlined]{algorithm2e}
\usepackage{makecell}

\title{\LARGE \bf
LLM-BT: Performing Robotic Adaptive Tasks based on Large Language Models and Behavior Trees
}

\author{Haotian Zhou$^{1\dag}$, Yunhan Lin$^{1\dag}$, Longwu Yan$^{1}$, Jihong Zhu$^{2}$ and Huasong Min$^{1}$* %  <-this % stops a space
\thanks{$^{\dag}$These authors contributed equally to this work. *Coresponding author: Huasong Min. This work is supported by the National Key R\&D Program of China (grant No.: 2022YFB4700400), National Natural Science Foundation of China (grant No.: 62073249), Key R\&D Program of Hubei Province (grant No.: 2023BBB011).}% <-this % stops a space
\thanks{$^{1}$Haotian Zhou, Yunhan Lin, Longwu Yan and Huasong Min are with Institute of Robotics and Intelligent Systems, Wuhan University of Science and Technology, Wuhan, China. {\tt\small zhtwust@163.com, yhlin@wust.edu.cn, longwuyan0621@163.com, mhuasong@wust.edu.cn.}}%
\thanks{$^{2}$Jihong Zhu is with the School of Physics, Engineering and Technology, University of York, the UK. {\tt\small jihong.zhu@york.ac.uk.}}%
}

\begin{document}

\maketitle
\thispagestyle{empty}
\pagestyle{empty}

%%%%%%%%%%%%%%%%%%%%%%%%%%%%%%%%%%%%%%%%%%%%%%%%%%%%%%%%%%%%%%%%%%%%%%%%%%%%%%%%
\begin{abstract}
Large Language Models (LLMs) have been widely utilized to perform complex robotic tasks. However, handling external disturbances during tasks is still an open challenge. This paper proposes a novel method to achieve robotic adaptive tasks based on LLMs and Behavior Trees (BTs). It utilizes ChatGPT to reason the descriptive steps of tasks. In order to enable ChatGPT to understand the environment, semantic maps are constructed by an object recognition algorithm. Then, we design a Parser module based on Bidirectional Encoder Representations from Transformers (BERT) to parse these steps into initial BTs. Subsequently, a BTs Update algorithm is proposed to expand the initial BTs dynamically to control robots to perform adaptive tasks. Different from other LLM-based methods for complex robotic tasks, our method outputs variable BTs that can add and execute new actions according to environmental changes, which is robust to external disturbances. Our method is validated with simulation in different practical scenarios.
\end{abstract}

%%%%%%%%%%%%%%%%%%%%%%%%%%%%%%%%%%%%%%%%%%%%%%%%%%%%%%%%%%%%%%%%%%%%%%%%%%%%%%%%
\section{Introduction}
Large Language Models (LLMs) \cite{ref_Brown} demonstrate powerful reasoning capabilities in robotics. By utilizing LLMs, robots can efficiently understand user intentions \cite{ref_Bucker} and deduce the workflow of tasks \cite{ref_Huang}. However, the application of LLMs in complex robotic tasks faces challenges. One challenge is that the descriptive steps generated by LLMs are not executable without knowledge of operational skills.

Recently, there are many methods for solving the grounding problems of LLMs in robotics. PaLM-E \cite{ref_Driess} generates control sentences according to multi-modal data. RT-2 \cite{ref_Brohan_2} directly infer instructions based on languages and images. ChatGPT for Robotics \cite{ref_Vemprala} needs the declaration of APIs for reasoning the actions of tasks. SayCan \cite{ref_Ahn} selects most suitable actions according to environmental information. VoxPoser \cite{ref_Huang_2} converts the observation space into a 3D value maps for generating trajectories. These methods are able to achieve robotic autonomous tasks and handle some partial disturbances. For example, when objects drop from the manipulator of robots, they can re-pick them.

However, they cannot handle external disturbances that require re-planning. For example, when an obstacle $O_{b}$ prevents the robot from placing object $O_{a}$ at position $P_{a}$, they cannot put down $O_{a}$, then move $O_{b}$ away, and finally pick $O_{a}$ and place it at $P_{a}$. This is because these methods are unable to add new actions with a higher executing priority at runtime.

Behavior Trees (BTs) \cite{ref_Colledanchise} were popular in video games \cite{ref_Rabin} and are widely used in robotics due to their unique properties of modularity and reactivity \cite{ref_Colledanchise_4}. BTs can react to environmental changes by frequent traversal (which calls tick) to activate actions \cite{ref_Colledanchise_3}. The synthesis of BTs \cite{ref_Colledanchise_2} allows new actions to be added with a higher executing priority in real time, which enables robots to deal with external disturbances. However, this kind of method needs to pre-define the goal of tasks.

\begin{figure}[htbp]
\centerline{\includegraphics[width=3.2in]{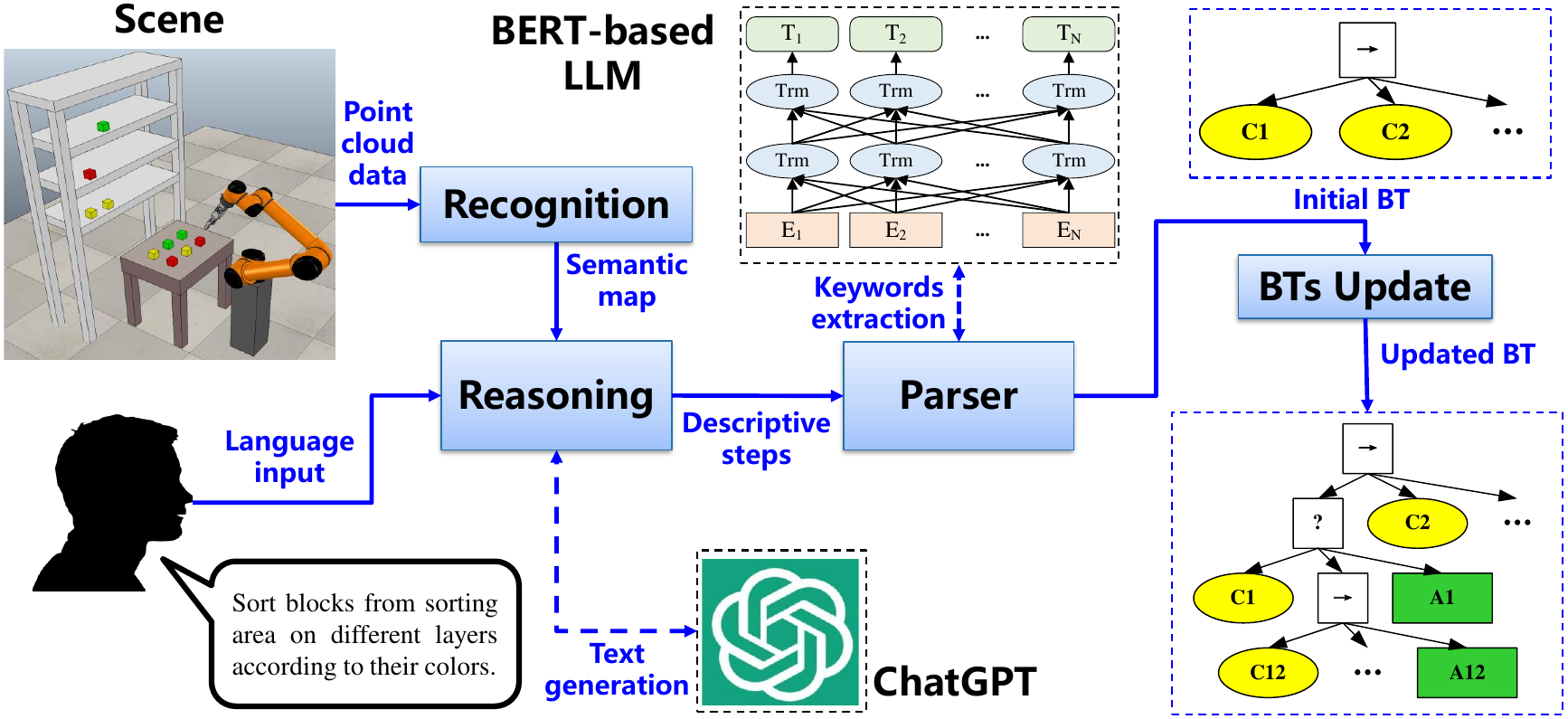}}
\caption{Overview of LLM-BT.}
\label{method_fig}
\end{figure}

In this paper, we utilize LLMs to construct initial BTs that include the goal of tasks and propose a BTs Update algorithm to expand the initial BTs. Fig. \ref{method_fig} shows the overview of our method which calls LLM-BT\footnote[1]{https://github.com/henryhaotian/LLM-BT}.

The Recognition module obtains information of objects in real-time scene to construct semantic maps. In the Reasoning module, ChatGPT \cite{ref_Liu} is adopted to deduce the descriptive steps of tasks based on the user inputs and semantic maps. Next, the Parser module utilizes a Bidirectional Encoder Representations from Transformers (BERT)-based \cite{ref_Devlin} LLM to extract keywords from the descriptive steps and then constructs initial BTs. Finally, a BTs Update algorithm is proposed to expand the initial BTs dynamically, which add and execute actions iteratively to satisfy the goal of tasks.

Compared to other LLM-based methods for complex robotic tasks, we have two advantages.

(1) Adaptability: LLM-BT outputs variable BTs that can add and execute new actions according to environmental changes, which is robust to external disturbances, as shown in Table \ref{comparison}.

(2) Modularity: LLM-BT consists of four modules, and each module can adopt advanced algorithms in the future. 
\begin{table}[h]
\centering
\caption{Comparison with LLM-based methods for complex robotic tasks.}
\label{comparison}
\begin{tabular}{ccc}
\hline
Methods & Input & Output \\
\hline
LLM-BT(Ours) & \makecell{language, \\ images} & variable BTs \\
\hline
PaLM-E \cite{ref_Driess} & \makecell{languages, \\ multi-modal data} & control sentences \\
\hline
RT-2 \cite{ref_Brohan_2} & \makecell{languages, \\ images} & low-level instructions \\
\hline
\makecell{ChatGPT \\ for Robotics \cite{ref_Vemprala}} & \makecell{languages, \\ declaration of APIs} & actions linked to APIs \\
\hline
SayCan \cite{ref_Ahn} & \makecell{languages, \\ images} & selected actions \\
\hline
VoxPoser \cite{ref_Huang_2} & \makecell{languages, \\ images} & motion trajectories \\
\hline
\end{tabular}
\end{table}

The main contributions of this paper are as follows:

(1) We propose a novel method to perform robotic adaptive tasks by constructing BTs automatically and expanding them dynamically. To the best of our knowledge, this is the first time that BTs have been generated through LLMs.

(2) We compare our method with advanced LLM-based methods and discuss the advantages and limitations of our method.

The rest of this paper is organized as follows. Section II is the related work. Section III introduces BTs briefly. Section IV illustrates the proposed method. Section V provides the experiments. Section VI is a discussion of our method and Section VII lists the conclusion and future work.

\section{Related work}
\subsection{LLM-based methods for complex robotic tasks}
PaLM-E \cite{ref_Driess} integrated an advanced LLM and a vision model for end-to-end training. It utilized text and multi-modal data (images, robot states, environmental information, etc.) instead of pure text as input, and outputted control sentences. 

RT-1 \cite{ref_Brohan_1} adopted imitation learning to acquire skills for robots. It first converted text instructions and images into tokens, and then used a Token Learner to compress these tokens in order to improve the model’s inference speed. Finally, the compressed tokens were transmitted into a Transformer for training. Subsequently, based on RT-1, RT-2 \cite{ref_Brohan_2} used a dataset of robot skills to fine-tune model, which significantly enhancing the capability of task generalization. 

ChatGPT for Robotics \cite{ref_Vemprala} created an advanced function library and then linked it to API of actual robotic platform. So, the robot can parse user intention and convert it into high-level function calls.

SayCan \cite{ref_Ahn} combined LLMs with Value Function. It used LLMs to output available high-level actions. Then, the value Function assigned scores to these actions based on environmental information. The scores represented the probability of these actions being executed respectively. Therefore, the action with a highest score would be executed.

VoxPoser \cite{ref_Huang_2} used LLMs and Vision-Language Models (VLMs) to convert the observation space into a 3D value maps, and then adopted mature motion planning algorithms to achieve adaptive tasks.

Differ from these methods, LLM-BT outputs variable BTs that can add new actions and assign their priorities based on environmental changes.

\subsection{Synthesis of BTs}
Excepts the method that synthesizes BTs based on \textit{Blended Reactive Planning and Acting} \cite{ref_Colledanchise_2}, advanced AI algorithms are utilized for the
acquisition of BTs in many approaches.

Scheper, et al \cite{ref_Scheper} employed genetic algorithms to select, crossover, and mutate nodes or subtrees in BTs. However, a complete BT must be manually constructed at first for subsequent optimization. French, et al \cite{ref_French} learned BTs from demonstration. They used Classification and Regression Tree algorithm to generate decision trees from human demonstrations and then converted the decision trees into BTs. Banerjee \cite{ref_Banerjee} learned BTs by reinforcement learning, which designed a conversion rule to transform the trained Q-tables into BTs.

The BTs Update algorithm in our method is related to \cite{ref_Colledanchise_2}. The difference is that we construct the initial BTs that include the goal of tasks by LLMs.

\section{Background:Behavior Trees}
A BT consists of control flow nodes, execution nodes and a root node. Control flow nodes traverse their child nodes based on some logic. Execution nodes are able to check conditions or perform actions. The execution of a BT starts from the root node, which ticks its child nodes. A ticked node returns a status of $Success$, $Faliure$ or $Running$ to its parents. \emph{Fallback} node, and \emph{Sequence} node are the commonly used control flow nodes.

\emph{Fallback} node: It ticks its child nodes from left to right. If a child node returns $Failure$, it ticks the next child node. If a child node returns $Success/Running$, it returns $Success/Running$ and stop ticking. Only all child nodes return $Failure$, does it return $Failure$. A `$?$' in box is used to represent a \emph{fallback} node.

\emph{Sequence} node: It ticks its child nodes from left to right. If a child node returns $Success$, it ticks the next child node. If a child node returns $Failure/Running$, it returns $Failure/Running$ and stop ticking. Only all child nodes return $Success$, does it return $Success$. A `$\rightarrow$' in box is used to represent a \emph{sequence} node.

Execution nodes are categorized as two kinds:

\emph{Action} node: a ticked \emph{action} node perform an action of robot. It returns $Success/Failure/Running$, indicating the action is finished/failed/being executed. A rectangle is used to represent an \emph{action} node.

\emph{Condition} node: a ticked \emph{condition} node checks whether a condition is satisfied. It returns $Success/Failure$, indicating the condition is true/false. An ellipse is used to represent a \emph{condition} node.

The execution mechanism of action nodes in BTs is similar to PDDL \cite{ref_Fox}. When an \emph{action} node in a BT needs to be executed, some \emph{condition} (known as pre-conditions) nodes must be satisfied firstly. Once an action node is completed, some other condition (known as post-conditions) nodes are then satisfied. The paradigm of an action node can be defined as Formula \ref{eq1}.
\begin{equation}
\begin{split}
& a = \{ \texttt{Pre}(a),\texttt{Post}(a) \} \\
& \texttt{Pre}(a) = \{ c^{pre}_{1},...,c^{pre}_{n} \} \\
& \texttt{Post}(a) = \{ c^{post}_{1},...,c^{post}_{m} \}
\label{eq1}
\end{split}
\end{equation}

Where $a$ represents an \emph{action} node, $c$ represents a \emph{condition} node. $\texttt{Post}(a)$ indicates the post-conditions of $a$, $\texttt{Pre}(a)$ indicates the pre-conditions of $a$.

\section{The Proposed Method}
In this section, we introduce the modules in LLM-BT, which are Recognition, Reasoning, Parser and BTs Update.

\subsection{Recognition}
\begin{figure}[htbp]
\centerline{\includegraphics[width=3.0in]{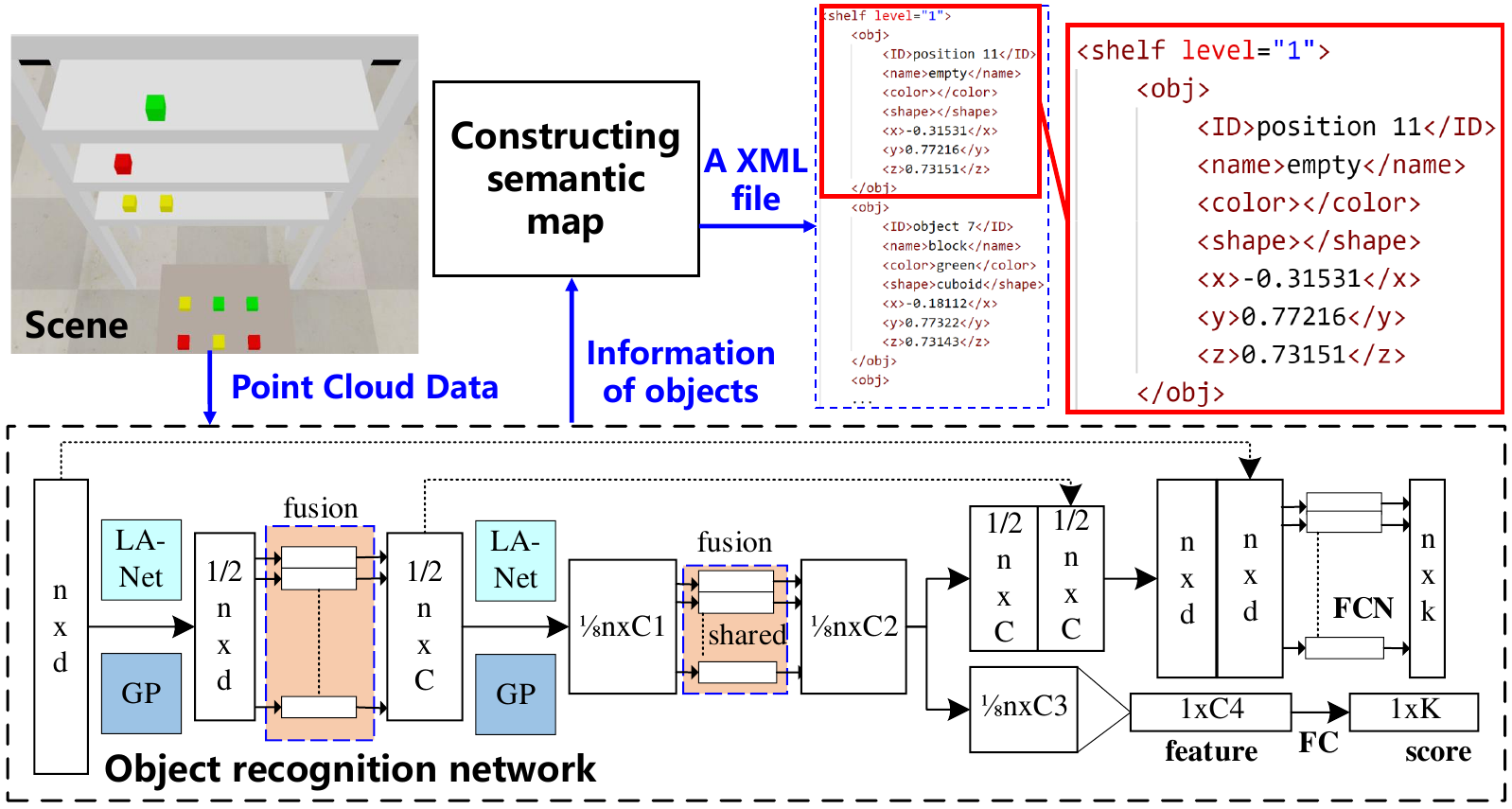}}
\caption{Pipeline of Recognition.}
\label{recognition_fig}
\end{figure}

Fig. \ref{recognition_fig} is the pipeline of Recognition. Firstly, point cloud data from the real-time scene is captured by a 3D camera. We proposed a 3D object recognition algorithm \cite{ref_Lin_3} to obtain information of objects, such as their ID, name, color, shape, and 3D spatial coordinates. Subsequently, this information is written in a semantic map as XML format.

We have chosen XML as the format for storing the semantic map because of its excellent extensibility, clear structure, and its ability to effectively represent relationships between different attributes.

\subsection{Reasoning}
\begin{figure}[htbp]
\centerline{\includegraphics[width=3.0in]{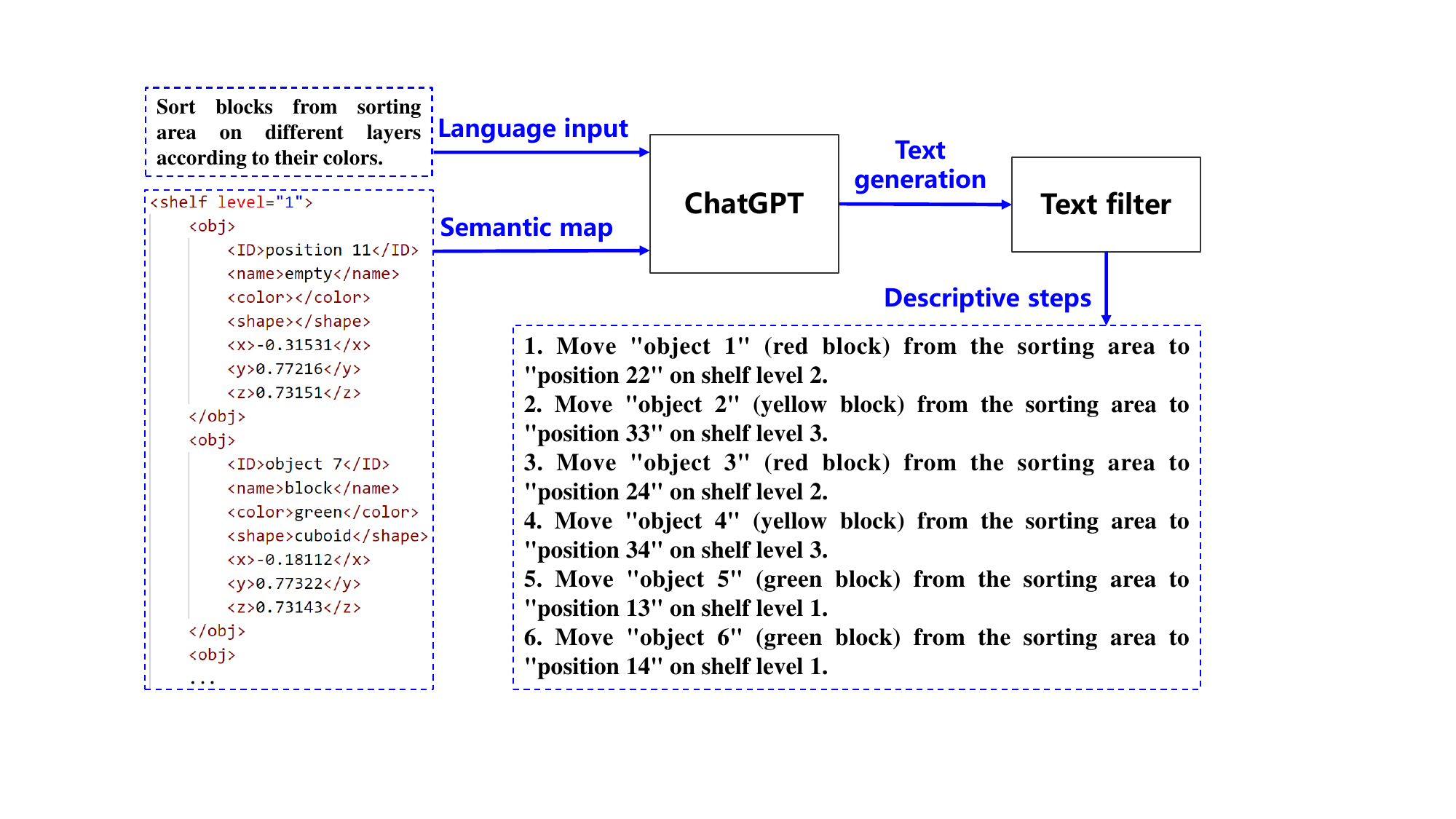}}
\caption{Pipeline of Reasoning.}
\label{reasoning_fig}
\end{figure}

In Reasoning module, we utilize the reasoning ability of ChatGPT \cite{ref_Liu} to understand the information from semantic map and user input. Due to the text generated by ChatGPT not only include descriptive steps, but also many explanatory or declarative statements, a text filter is designed to prevent interference with subsequent keyword extraction. Fig. \ref{reasoning_fig} shows the pipeline of Reasoning.

\subsection{Parser}
\begin{figure}[htbp]
\centerline{\includegraphics[width=2.7in]{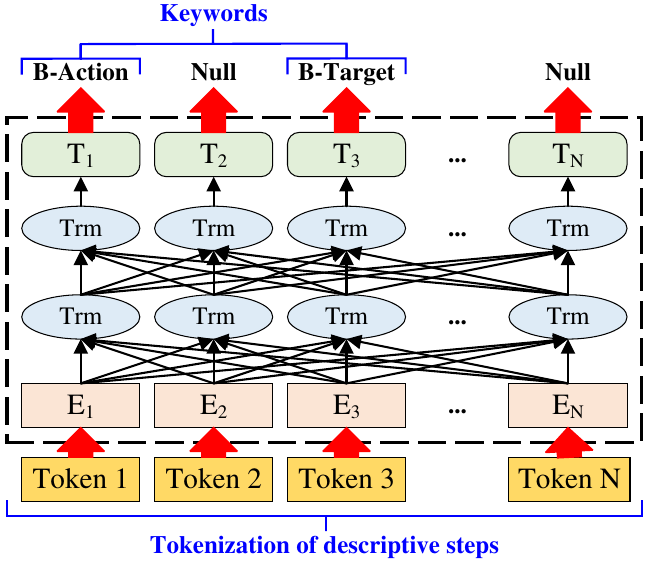}}
\caption{Pipeline of \textit{keywords extraction}.}
\label{bert_fig}
\end{figure}

Parser module consists of two parts, which are \textit{keywords extraction} and \textit{keywords parsing}. The \textit{keywords extraction} uses a BERT-based \cite{ref_Devlin} LLM to extract keywords. Fig \ref{bert_fig} shows the pipeline of \textit{keywords extraction}. The descriptive steps are first split by a tokenizer into a set of tokens as inputs of the LLM. Then, the tokens which are labeled are regarded as keywords.

During pre-training, we designed six types of labels in datasets, which are 'B-Action', 'B-Target', 'I-Target', 'B-Destination', 'I-Destination', 'B-Location' and 'I-Location'. 'B-Action' represents a token is the name of an action. 'B-Target' or 'I-Target' represents a token is the name of an operating objective. 'B' means the first word of the name and 'I-Target' means the subsequent word of the name. Similarly, 'B-Destination' or 'I-Destination' represents a token is the name of a destination for operating. 'B-Location' or 'I-Location' represents a token is the name of a location.

\begin{figure}[htbp]
\centerline{\includegraphics[width=3.2in]{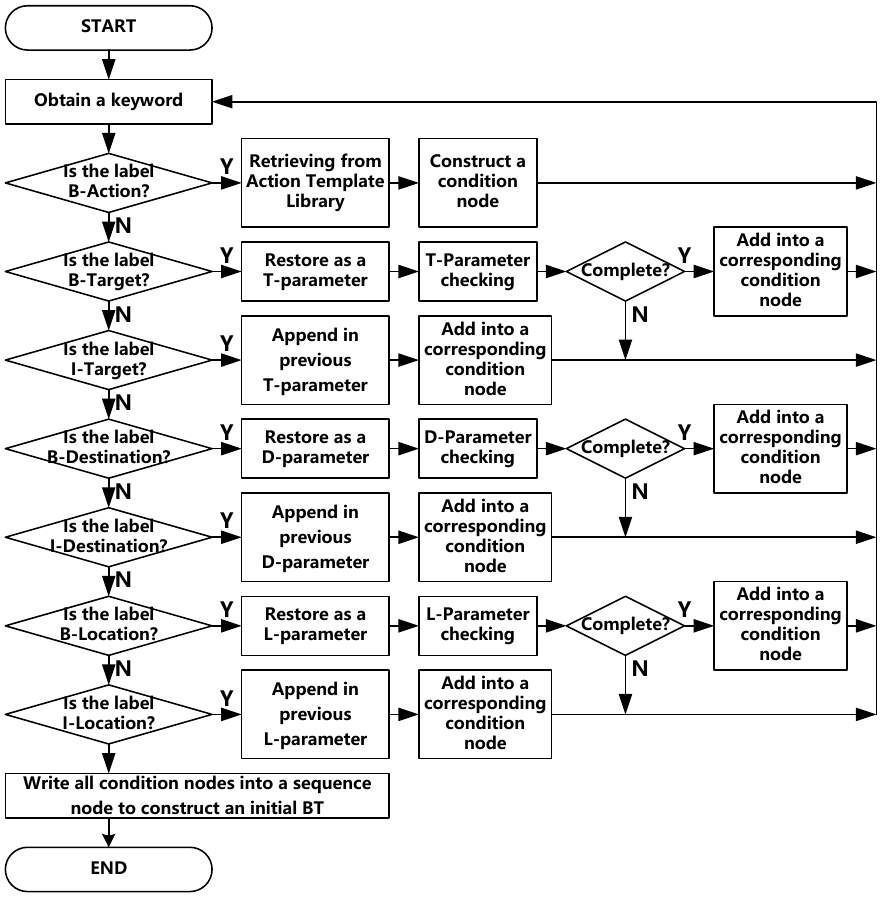}}
\caption{The process of \textit{keywords parsing}.}
\label{parser_fig}
\end{figure}

Subsequently, these keywords are parsed. Fig. \ref{parser_fig} shows the process of \textit{keywords parsing} based on our previous automatic programming \cite{ref_Lin_2}.

The overall parsing is based on a “if-else” structure. Each keyword will be read in turn. When the label of a keyword $w$ is 'B-Action', $w$ represents an action $a$. Then, ATL will retrieve an appropriate condition node. The action templates in ATL are designed based on Formula \ref{eq1}. According to $a$, a condition node $c$ that satisfies $c \in Post(a)$ will be found in ATL.

Then, a condition node $c \in Post(a)$ which can be satisfied after completing $a$ will be found from an Action Template Library (ATL). The action templates in ATL are designed based on Formula \ref{eq1}. So the

When $w$ belongs to 'B-Target',$w$ is a T-parameter (a parameter of the operating objective). Then, $w$ needs to be checked for the integrity of a T-parameter. This is because some parameters include multiple keywords, such as 'position 1' or 'potato chips'. If $w$ is complete, it will be added into a corresponding condition node. otherwise, a new keyword is read for a new round of processing.

When $w$ belongs to 'I-Target', $w$ is a part of previous T-parameter. Then, $w$ is integrated into a T-parameter. And this T-parameter will be added to a corresponding condition node. 

Similarly, D-parameter (a parameter of the destination for operating) and L-parameter (a parameter of the location) are obtained and added to corresponding condition nodes.

\begin{figure}[htbp]
\centerline{\includegraphics[width=3.2in]{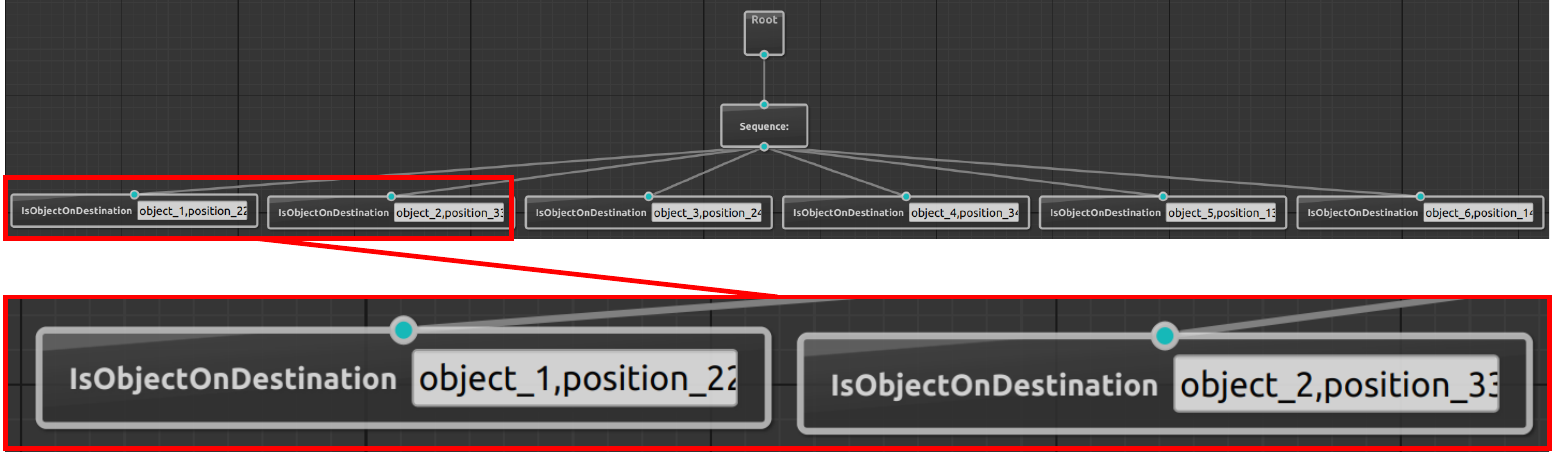}}
\caption{An initial BT constructed by Parser.}
\label{initial_BT_fig}
\end{figure}

When all keywords have been traversed, it writes all conditions node into a sequence node to construct an initial BT which represents the goal of a task. Fig. \ref{initial_BT_fig} shows the initial BT constructed by Parser based on the descriptive steps in Fig. \ref{reasoning_fig}.

Due to the manipulator is fixed in this scene (Fig. \ref{method_fig}), only T-parameters and D-parameters are useful in condition nodes. For mobile manipulation robots, L-parameters are required (refer to experiments of household service in section V.B the home service experiment).

\subsection{BTs Update}
\subsubsection{The process of BTs Update}
BTs Update mainly includes two functions: \texttt{Expand} and \texttt{Insert}. Algorithm \ref{al1} shows its pseudo-code. Where $A$ represents a set of \emph{action} nodes and $C$ represents a set of \emph{condition} nodes. $\mathcal{T}$ denotes a BT or a subtree and $\Gamma$ represents a set of subtrees. $R$ is the returned status of a BT.
\begin{algorithm}
\caption{BTs Update}
\label{al1}
\While {$R \neq Success$}
{
    $R\leftarrow \texttt{Execute}(\mathcal{T})$\;
    \If {$R = Failure$}
    {
        $c_{f}\leftarrow \texttt{GetFailedNode}(\mathcal{T})$\;
        $\mathcal{T}_{exp},\texttt{Pre}(a) \leftarrow \texttt{Expand}(c_{f})$\;
        $\mathcal{T}\leftarrow \texttt{Insert}(\mathcal{T},c_{f},\mathcal{T}_{exp},\texttt{Pre}(a))$;
    }
}
\SetKwFunction{FMain}{$\texttt{Expand}$}
\SetKwProg{Fn}{Function}{:}{}
\Fn{\FMain{$c_{f}$}}
{
    $\mathcal{T}_{exp} \leftarrow \emptyset$\;
    $\Gamma \leftarrow \emptyset$\;
    \For {$a\ \KwTo\ A_{tp}$}
    {
        \If {$c_{f} \in \texttt{Post}(a)$}
        {
            $\mathcal{T}_{st} \leftarrow \textsf{SequenceNode}(\texttt{Pre}(a),a)$\;
            $\Gamma \leftarrow \mathcal{T}_{st}$\;
        }
    }
    $\mathcal{T}_{exp} \leftarrow \textsf{FallbackNode}(c_{f},\Gamma)$\;
    \Return $\mathcal{T}_{exp},\texttt{Pre}(a)$\;
}
\SetKwFunction{FMain}{$\texttt{Insert}$}
\SetKwProg{Fn}{Function}{:}{}
\Fn{\FMain{$\mathcal{T},c_{f},\mathcal{T}_{exp},\texttt{Pre}(a)$}}
{
$C_{h} \leftarrow \texttt{HigherPriorityNodes}(\mathcal{T}, c_{f})$\;
$C_{a} \leftarrow \texttt{Pre}(a)$\;
$C_{t} \leftarrow \texttt{Type}(C_{h}) \cap \texttt{Type}(C_{a})$\;
\eIf {$C_{t} = \emptyset$}
{
    $\mathcal{T}\leftarrow \texttt{Replace}(\mathcal{T},c_{f},\mathcal{T}_{exp})$;
}
{
    $\mathcal{T} \leftarrow \texttt{Add}(\mathcal{T},C_{t},\mathcal{T}_{exp})$\;
}
\Return $\mathcal{T}$\;
}
\end{algorithm}

At first, an initial BT is executed. When $\mathcal{T}$ returns $Failure$ after being executed, the \emph{condition} node $c_{f}$ that leads to the failure of $\mathcal{T}$ will be obtained by \texttt{GetFailedNode} (lines 3-4 of Algorithm \ref{al1}). Then, \texttt{Expand} constructs an expanded subtree $\mathcal{T}_{exp}$ and insert it into $\mathcal{T}$(lines 5-6). The purpose of \texttt{Expand} is to change $c_{f}$ from $Failure$ to $Success$ by performing new \emph{action} nodes.

If $\mathcal{T}$ still returns $Failure$, $\mathcal{T}$ will be expanded again. Therefore, BTs Update incrementally expands $\mathcal{T}$ to satisfy each failed node in $\mathcal{T}$.

\subsubsection{Expand}
\texttt{Expand} searches for appropriate actions in ATL $A_{tp}$. When there is an \emph{action} node $a$ whose post-conditions contains $c_{f}$, it means that the completion of $a$ will change $c_{f}$ from $Failure$ to $Success$ (lines 10-11 of Algorithm \ref{al1}). 

Then, a subtree $\mathcal{T}_{st}$ will be constructed to ensure the pre-conditions of $a$ are satisfied before executing $a$. Next, $\mathcal{T}_{st}$ is collected by $\Gamma$ (lines 12-13 of Algorithm \ref{al1}). Finally, a \emph{fallback} node will be set as the parent of $c_{f}$ and $\Gamma$ to construct a expanded subtree (lines 14-15 of Algorithm \ref{al1}). 

Here, $\textsf{SequenceNode}$ constructs a subtree with a \emph{sequence} node as its top node. The children of it can be a node ($a$ or $c$) or a subtree ($\mathcal{T}$). Similarly, $\textsf{FallbackNode}$ constructs a subtree with a \emph{fallback} node as its top node.

\begin{figure}[htbp]
\centerline{\includegraphics[width=2.8in]{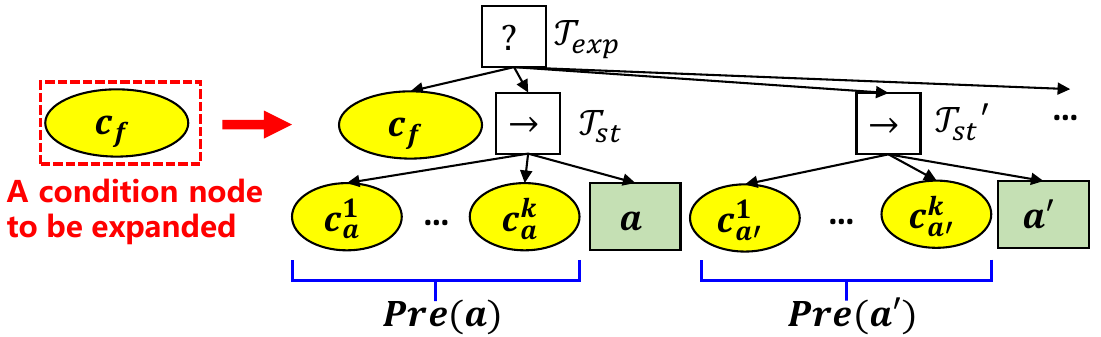}}
\caption{The process of \texttt{Expand}.}
\label{expand_fig}
\end{figure}

When $a$ is completed, $c_{f}$ may be satisfied. And when $c_{f}$ returns $Success$, other nodes in $\mathcal{T}_{exp}$ will no longer be traversed. Fig. \ref{expand_fig} is the process of \texttt{Expand}, it gives a comprehensive solution for $c_{f}$.

\subsubsection{Insert}
\texttt{Insert} analyzes the conflict between $\mathcal{T}$ and $\mathcal{T}_{exp}$ and then inserts $\mathcal{T}_{exp}$ into a appropriate position in $\mathcal{T}$. The \emph{condition} nodes in $\mathcal{T}$ with higher priority than $c_{f}$ (rank before $c_{f}$) are collected by \texttt{HigherPriorityNodes}. If set $C_{h}$ and set $C_{a}$ do not contain nodes of the same type, it denotes that there is no conflict between $\mathcal{T}$ and $\mathcal{T}_{exp}$. Then, $c_{f}$ is replaced by $\mathcal{T}_{exp}$ in $\mathcal{T}$ (lines 19-21 of Algorithm \ref{al1}).

\begin{figure}[htbp]
\centerline{\includegraphics[width=2.8in]{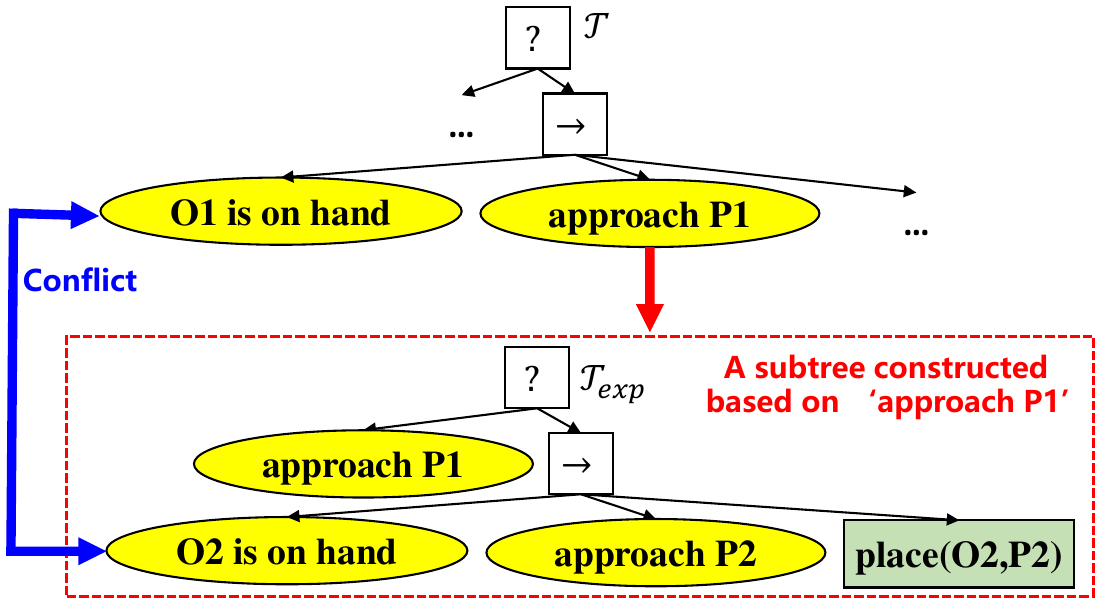}}
\caption{An example of conflict.}
\label{conflict_fig}
\end{figure}

If $C_{h}$ and $C_{a}$ contain nodes of the same type, replacing $c_{f}$ by $\mathcal{T}_{exp}$ directly will cause a conflict. This is because the purpose of $\mathcal{T}_{exp}$ is to perform a new action. If there is a node in $C_{h}$ returns $Success$, a node in $C_{a}$ will never be satisfied. Fig. \ref{conflict_fig} is an example of conflict.
\begin{figure}[htbp]
\centerline{\includegraphics[width=2.8in]{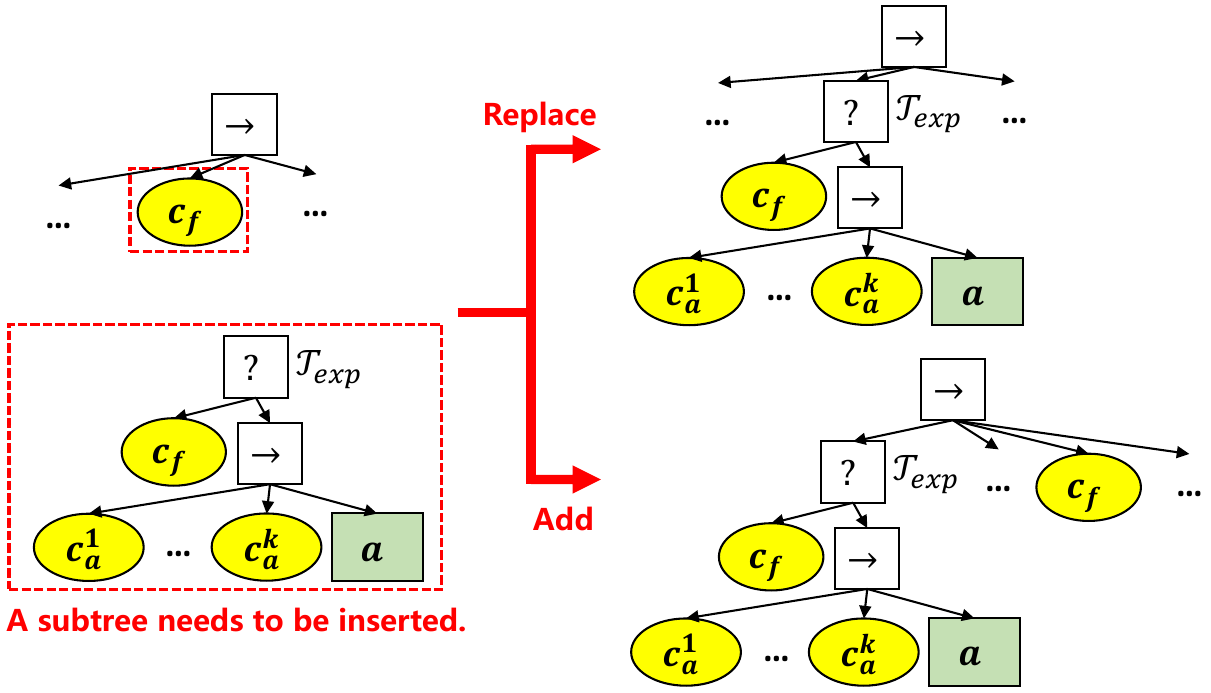}}
\caption{The process of \texttt{Insert}.}
\label{insert_fig}
\end{figure}

At this time, $\mathcal{T}_{exp}$ is added before all nodes in $C_{t}$ for raising the priority of $T_{exp}$ (lines 22-23 of Algorithm \ref{al1}). Fig. \ref{insert_fig} gives the process of \texttt{Insert}.

\section{Experiments}
We conduct two kinds of experiments which are Cargo sorting and Household service. In Cargo sorting, a fixed manipulator needs to sort cargo according to the user’s request. In Household service, a mobile manipulation robot needs to bring objects required for user.
\subsection{Cargo sorting}
There are 5 different cases designed in Cargo sorting. Each case invites 20 non-professional users to communicate with the robot for informing their request of sorting tasks (similar to Fig. \ref{method_fig}). Meanwhile, there is a random external disturbance (such as an object drops from the manipulator or is blocked by an obstacle) deployed every time the robot works.

To control the variables, the success rates of actions and 3D recognition are not taken into account.
\begin{table}[htbp]
\centering
\caption{The results of Cargo sorting.}
\label{sorting_res}
\begin{tabular}{ccccc}
\hline
Cases & \makecell{Overall \\ (S/ALL)} & \makecell{Reasoning \\ (S/ALL)} & \makecell{Parser \\ (S/ALL)} &  \makecell{BTs Update \\ (S/ALL)} \\
\hline
Case 1 & 18/20 & 18/20 & 18/18 & 18/18 \\
Case 2 & 17/20 & 18/20 & 17/18 & 17/17 \\
Case 3 & 16/20 & 17/20 & 17/17 & 16/17 \\
Case 4 & 18/20 & 18/20 & 18/18 & 18/18 \\
Case 5 & 18/20 & 19/20 & 18/19 & 18/18 \\
\hline
\end{tabular}
\end{table}

Table \ref{sorting_res} shows the results. Where 'Overall' represents the overall success rate of Cargo sorting. 'S/ALL' represents the number of successes out of 20 tests. A success is considered respectively when 'Reasoning' deduces a set of correct descriptive steps, 'Parser' constructs a correct initial BT, and 'BTs Update' returns $Success$.
\begin{figure}[htbp]
\centerline{\includegraphics[width=3.2in]{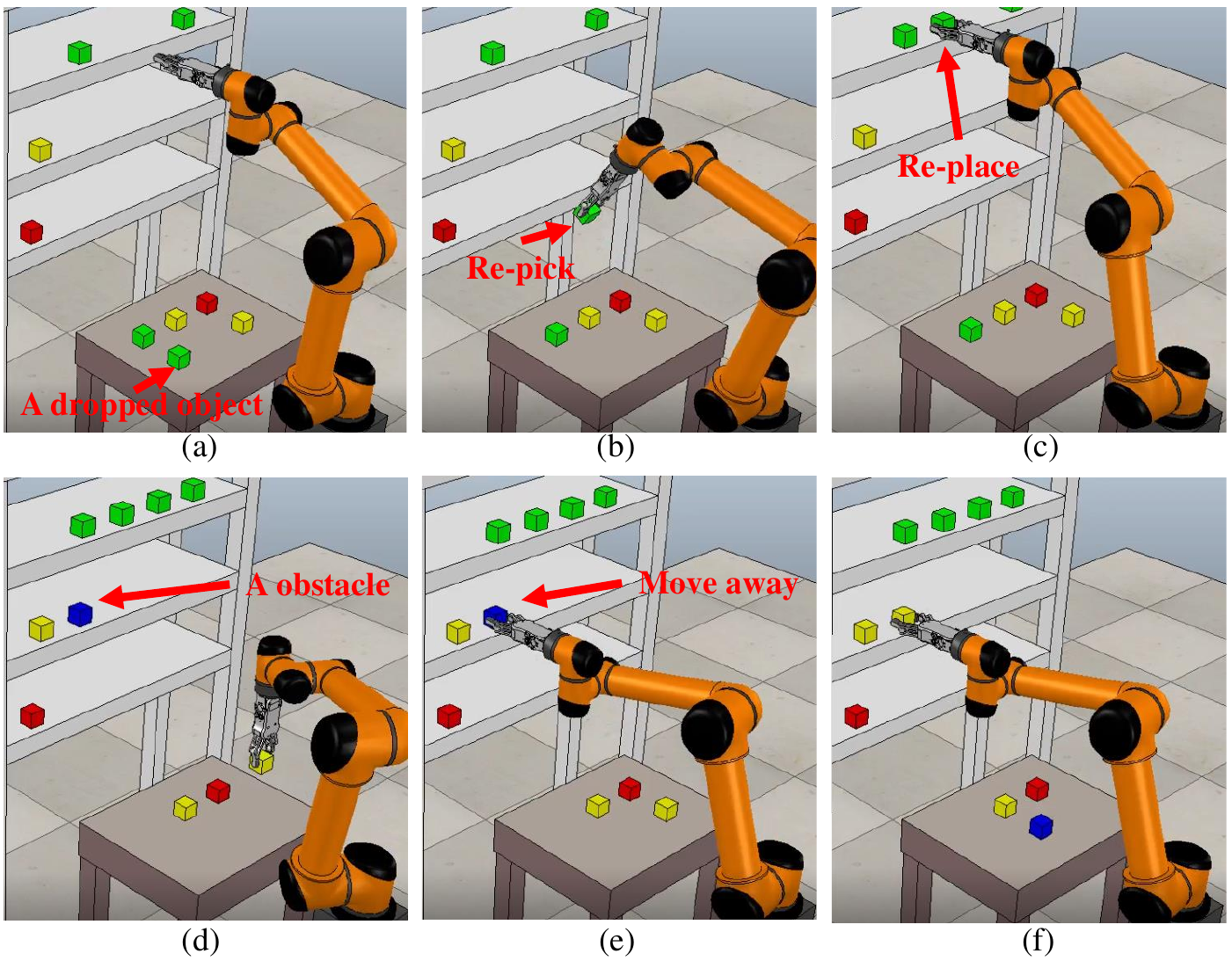}}
\caption{Dealing with external disturbances during sorting tasks.}
\label{cargo_sorting_fig}
\end{figure}

The success rate of Cargo sorting was about 85\%. In these 5 cases, there were 10 failures caused by unclear requirements expressed by users, which led to the incorrect descriptive steps deduced by Reasoning module. There was 1 failure caused by a unencountered keyword, so that the Parser module constructed an incorrect initial BT. There was 1 failure caused by an unsolvable external disturbance, which BTs Update returned $Failure$.

Apart from these failures, the robot was able to achieves robotic adaptive tasks. As shown in Fig. \ref{cargo_sorting_fig}, the robot could pick up an object again when the object dropped from the end effector (Fig. \ref{cargo_sorting_fig}(a)(b)(c)). It could move an obstacle away when it hindered a picking or placing action (Fig. \ref{cargo_sorting_fig}(d)(e)(f)).

\subsection{Household service}
There are also 5 different cases designed in Household service. Each case invites 20 non-professional users to communicate with the robot for informing their request. There is also a random external disturbance deployed every time the robot works.
\begin{table}[htbp]
\centering
\caption{The results of Household service.}
\label{household_res}
\begin{tabular}{ccccc}
\hline
Cases & \makecell{Overall \\ (S/ALL)} & \makecell{Reasoning \\ (S/ALL)} & \makecell{Parser \\ (S/ALL)} &  \makecell{BTs Update \\ (S/ALL)} \\
\hline
Case 1 & 15/20 & 17/20 & 16/17 & 15/15 \\
Case 2 & 14/20 & 16/20 & 14/16 & 14/14 \\
Case 3 & 15/20 & 16/20 & 15/16 & 15/15 \\
Case 4 & 16/20 & 18/20 & 16/18 & 16/16 \\
Case 5 & 17/20 & 17/20 & 17/17 & 17/17 \\
\hline
\end{tabular}
\end{table}

\begin{figure}[htbp]
\centerline{\includegraphics[width=2.8in]{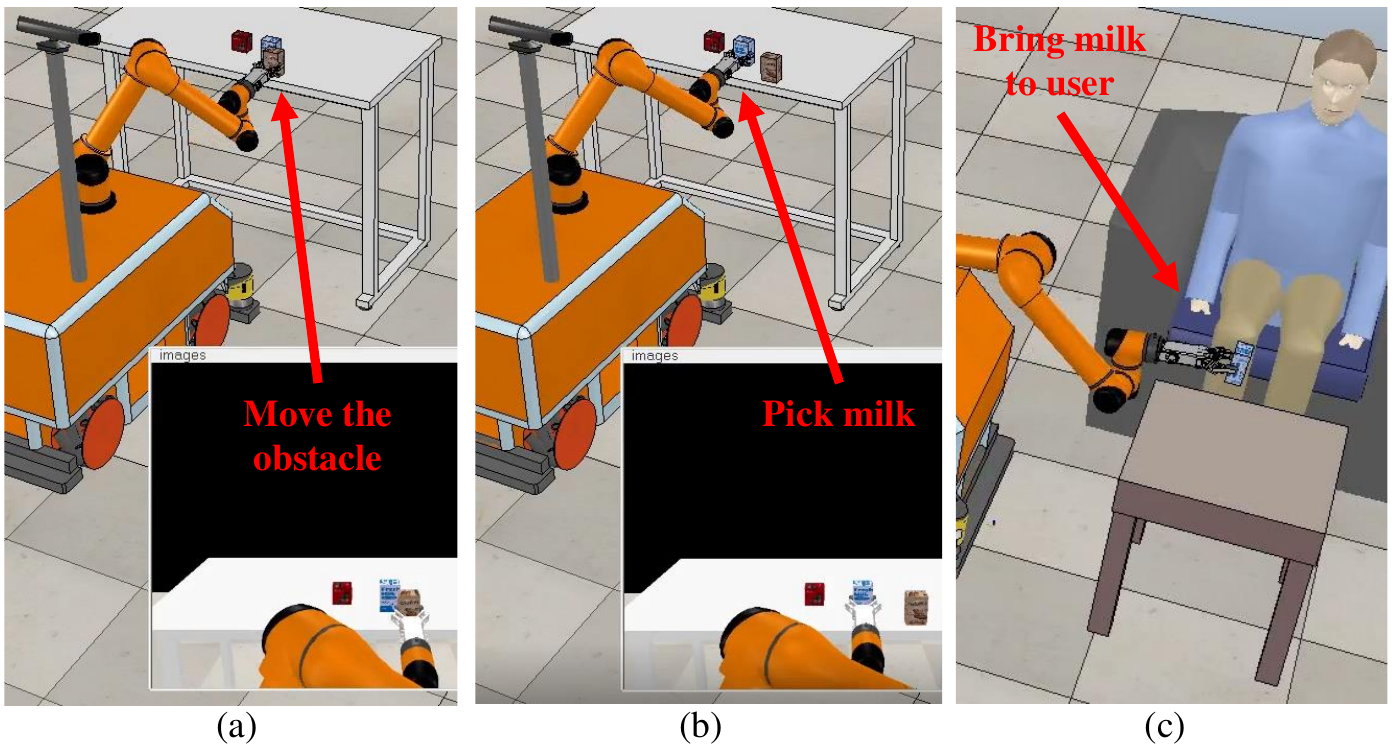}}
\caption{Dealing with external disturbances during household service. (a) the robot firstly moved the obstacle away from the milk, (b) picked the milk, (c) brought it to user.}
\label{household_service_fig}
\end{figure}

Table \ref{household_res} shows the results. The success rate of Household service decreased to about 75\%. With the increasing complexity of environment, user’s unclear expressions could lead to a higher probability of reasoning failure (16 out of 100 tests). The complicated environment also led to the generation of novel descriptive steps, which caused failure of Parser module (6 out of 84 tests). Therefore, expanding the dataset of BERT-based LLM is an important part for improving LLM-BT. Apart from these failures, the robot was able to achieves robotic adaptive tasks, as shown in Fig. \ref{household_service_fig}

Consequently, the results of these two experiments demonstrate the feasibility of our method. We demonstrate the video\footnote[2]{https://youtu.be/TkIWb53IEtQ} of Cargo sorting and Household service.

\section{Discussion}
\subsection{Advantages}
Compared to other LLM-based methods, the notable advantage of our method is adaptability. LLM-BT is capable of adding new actions through ATL based on environmental changes and assigning appropriate executing priorities to these actions.

The environmental changes may result in different operations for the same task. We consider that an effective method is to utilize LLM to deduce the step of a task. Then, paring it into a set of conditions that need to be satisfied. And finally using BTs Update algorithm to execute appropriate actions based on the current environment.

\subsection{Limitations}
Although LLM-BT demonstrates adaptability, there are two limitations in our method.

(1) Relatively weak understanding of scene: The system is unable to learn new knowledge from the semantic map. For example, in cargo sorting experiments, the system only places each kind of object on the empty position of corresponding layer, without arranging them in a specific order. This is because it cannot infer the relative positions of objects based on their 3D spatial coordinates.

(2) Manually construct ATL: the ATL is a crucial component for Parser module and \texttt{Expansion} in BTs Update. The action templates in the ATL are added manually by experts or engineers. When the system needs to add a new kind of action node, a corresponding action template also needs to be added.

\section{Conclusion and future work}
In this paper, LLM-BT is proposed. It utilizes LLMs to construct initial BTs automatically, and then uses BTs Update algorithm to expand them dynamically. The experiments demonstrate that LLM-BT enables robots to perform robotic adaptive tasks.

In the future work, we will focus on improving two parts of LLM-BT. The first part is to improve the parsing accuracy of keywords in the operation parsing module by utilizing an advanced deep learning model. The second part is to apply robotic manipulation algorithms for deformable objects to conduct complicated experiments.

\end{document}